# Chatting with Bots: AI, Speech Acts, and the Edge of Assertion

**Iwan Williams (ORCID: 0000-0003-0582-0983)** [1,2]
iwan.williams1@monash.edu

**Tim Bayne (ORCID: 0000-0001-8591-7907)** [1,2,3]
timothy.bayne@monash.edu

[1]Department of Philosophy, School of Philosophical, Historical and International Studies, Monash University, Melbourne, Victoria, Australia.

[2]Monash Centre for Consciousness & Contemplative Studies (M3CS), Monash University, Melbourne, Victoria, Australia.

[3]Brain, Mind and Consciousness Program, Canadian Institute for Advanced Research (CIFAR), Toronto, Ontario, Canada.

## Abstract

This paper addresses the question of whether large language model-powered chatbots are capable of assertion. According to what we call the *Thesis of Chatbot Assertion* (TCA), chatbots are the kinds of things that can assert, and at least some of the output produced by current-generation chatbots qualifies as assertion. We provide some motivation for TCA, arguing that it ought to be taken seriously and not simply dismissed. We also review recent objections to TCA, arguing that these objections are weighty. We thus confront the following dilemma: how can we do justice to both the considerations for and against TCA? We consider two influential responses to this dilemma—the first appeals to the notion of proxy-assertion; the second appeals to fictionalism—and argue that neither is satisfactory. Instead, reflecting on the ontogenesis of assertion, we argue that we need to make space for a category of proto-assertion. We then apply the category of proto-assertion to chatbots, arguing that treating chatbots as proto-assertors provides a satisfactory resolution to the dilemma of chatbot assertion.

# 1. Introduction

Chatbots, and the large language models (LLMs) that serve as their engines, have become a central part of social, intellectual and business life. Students use them to construct essays, academics use them to draft grant applications, real estate agents use them to write advertising copy, magazine editors use them to generate content, and many millions of people relate to them as friends and romantic partners. What, exactly, is the nature of our interactions with LLM-powered chatbots?

These interactions are routinely described as 'conversations'. There is, however, a real question as to whether such talk should be taken literally, or whether we should treat 'chatbot conversation' on a par with other applications of psychological terminology ('thinks', 'wants', etc.) to machines—useful, but not strictly speaking true. Indeed, many would argue that our interactions with chatbots are no more genuine conversations than monopoly dollars are legal tender, and that interacting with a chatbot involves nothing more than simulating a conversation.[1]

This paper examines the question of chatbot conversation by focusing on the case of assertion. Asserting (near synonyms are *stating*, *affirming*, or *claiming*) is one of the many things that we do with language—it is a speech act. Our assertions range from the mundane ("nice weather today", uttered at a bus stop) to the consequential ("it's cancer", delivered by a doctor). According to what we call the *Thesis of Chatbot Assertion* (TCA), LLM-powered chatbots are also the kinds of things that can assert, and at least some of the output produced by current-generation chatbots qualifies as assertion. Showing that TCA is true wouldn't show that chatbots are capable of full the range of speech acts that characterize human conversation, but—in addition to being an interesting result in its own right—it would go a long way towards vindicating the claim that chatbots can be conversationalists. And of course, if TCA turns out be false then the claim that chatbots are genuine conversationalists would have little plausibility, for assertion is a—arguably *the*—central speech act.

---

[1] This appears to be the position of IBM: "A chatbot is a computer program that simulates human conversation with an end user."
https://www.ibm.com/topics/chatbots#:~:text=A%20chatbot%20is%20a%20computer,and%20automate%20responses%20to%20them.

This paper unfolds as follows. Section 2 provides some motivation for TCA, arguing that it ought to be taken seriously and not simply dismissed. Section 3 considers recent objections to TCA, arguing that these objections are weighty. This leads us to something of a dilemma: how can we do justice to both the considerations for and against TCA? Section 4 considers two responses to this question: the first appeals to the notion of proxy-assertion; the second appeals to fictionalism. Reflecting on the ontogenesis of assertion, Section 5 argues that we need to make space for a category of proto-assertion. Section 6 then applies the category of proto-assertion to chatbots, arguing that we ought to treat at least certain kinds of chatbots as 'edge-cases' when it comes to the capacity for assertion.

## 2. Motivating the Thesis of Chatbot Assertion (TCA)

Ever since the development of ELIZA in the 1960s, we have been familiar with chatbots—computer programs that are able to produce natural language outputs on the model of ordinary human interaction. Although there is no serious reason to think that ELIZA is capable of assertion, things are notably different when it comes to current generation chatbots that are underpinned by LLMs (such as Claude or ChatGPT). Henceforth, by 'chatbots' we mean those that are powered by LLMs.

There are three reasons to take seriously TCA. First, not only are some chatbot outputs sources of information, being informative is arguably part of their proper *function* (Butlin, 2023; Butlin & Viebahn, forthcoming; Coelho Mollo & Millière, 2023). As Butlin & Viebahn (forthcoming) have argued, although the outputs of basic pretrained LLMs (those simply trained on next-token-prediction) might function merely to be statistically probable, models that have been fine-tuned in the appropriate way may acquire the function of outputting sentences that are true or informative. In their words, the LLMs have "descriptive functions" (p. 3). In any case, chatbots are used in many of the ways that they are only *because* their outputs are sufficiently informative sufficiently often. The fact that they are sufficiently informative plays a role in explaining their (continued) existence. This fits with many orthodox accounts of assertion according to which assertions "aim at truth" (Dummett, 1973; Marsili, 2018), have a "word-to-world" direction of fit (Searle, 1976), "present a proposition as true" (Wright 1992, p. 34; Adler 2002, p. 274) or have the proper function of inducing true beliefs in hearers (Graham, 2018; Simion & Kelp, 2018).

A second motivation for TCA concerns the contrast between a chatbot's different 'modes'. In addition to (apparently) asserting, chatbots appear to do other things with text, such as ask *questions*, give *recommendations* and issue *warnings*, engage in *play* and *pretence*, *quote* the speech of others, and much more. It is natural to describe these changes in mode in terms of changes in the *illocutionary force* of their linguistic outputs, that is, to treat them as involving different speech acts (including assertion). Certainly, we seem to do so implicitly: Researchers who worry about language models "hallucinating" or (better) "confabulating"—that is, outputting falsehoods that are presented as facts—presuppose the notion that some model outputs (but not others) *present things as facts* (see, e.g. Evans et al. 2021; Glaese et al., 2022). Correspondingly, we tend to be less concerned when a chatbot generates false statements in response to, say, a prompt that begins "Write a short fictional story about…" (Evans et al. 2021, p. 47).

A third motivation for TCA concerns the behavioural profile of chatbots. Chatbots do not merely generate assertion-like sentences in isolation, they display complex and flexible linguistic behaviour that closely resembles the behavioural profile of asserters. For instance, adult human assertions typically come with the expectation that the speaker will attempt to *defend* their assertion against reasonable challenges, to *explain* their reasons for thinking that the asserted statement is true, to not blatantly *contradict* their assertion in the next moment, and to *retract* their assertion if it is shown to be unsupported (Hamblin 1970; Brandom 1994; Marsili, 2024). While far from perfect, advanced chatbots meet all of these expectations to an impressive degree. The point of this observation is not to make a naïve inference from the behaviour of chatbots to a claim about their status as linguistic agents. The point is simply that the behavioural dispositions of current chatbots, which far outstrip the dispositions of simpler systems (such as thermometers, clocks and prerecorded messages), make a much stronger *prima facie* case for their being asserters.

These considerations obviously don't prove TCA, but they do provide significant motivation for it. TCA needs to be taken seriously, and not dismissed as misguided anthropomorphism. Taking it seriously, of course, requires reckoning with objections to it. We turn now to that task.

## 3. The Case Against Chatbot Assertion

This section presents four of the most prominent objections to TCA, leaving the task of evaluating them for later (Section 6).

### 3.1 The understanding objection

One objection to TCA focuses on the fact that understanding is a precondition of assertion. If making an assertion requires understanding the sentences one utters, and chatbots fail to understand their outputs, then it follows that TCA is false.

Much ink has been spilt on the question of whether chatbots—or artificial agents more generally—might have semantic understanding (Bender & Koller, 2020; Butlin, 2023; Mitchell & Krakauer, 2022). Famously, Searle's Chinese Room argument claims to show that mere computational processes don't suffice for understanding, and that genuine understanding requires consciousness. We set Searle's challenge to one side, assuming that anybody who takes seriously the possibility of AI assertion will have been persuaded by one of the many responses that have been made to Searle (see Preston & Bishop 2002).

However, even those who are willing to ascribe semantic understanding to certain types of AI agents might be reluctant to accept that contemporary chatbots understand what they 'say' given that their world is purely text-based. How, one might ask, could a purely disembodied system understand the full range of concepts that characterize human cognition, such as RED, FEATHERY, SAD, ANGRY, LOVE, BEAUTY and so on? The worry is that chatbot understanding could be at most a kind of syntactic, or formal linguistic competence—it couldn't reach beyond text and include a grasp of non-linguistic reality (Bender & Koller, 2020; Lake & Murphy, 2021; Mahowald et al., 2023). But if chatbots don't understand their outputs, then—so the understanding objection goes—those outputs fail to qualify as assertions.

**3.2 The attitudes objection**

An assertion isn't merely an utterance that is produced with understanding. After all, one can meaningfully say 'N'djamena is the capital of Chad' in order to improve one's pronunciation or to test a microphone, but neither of those actions qualify as assertions. What more does genuine assertion require? According to many accounts, part of what is required is the capacity to have certain *attitudes* to the contents of one's utterances. Thus, as has often been noted (Arora, 2024; Mallory, 2023; van Woudenberg et al. 2024; Connolly forthcoming), orthodox treatments of assertion place mental state demands that chatbots don't appear to meet.

Some early accounts held that a speaker must believe *p* in order to assert *p*. This is likely too strong, as *insincere* assertions can occur in the absence of a belief. Nonetheless, it seems plausible to hold that one must at least have a *capacity* to believe or disbelieve the propositions expressed by one's utterances, and some have appealed to this view to mount a challenge against TCA (Mallory 2023, p. 1085). The spirit of this objection is captured in a passage from Bernard Williams:

> when I said this machine made assertions, I should have actually put that in heavy scare-quotes; 'assertion' itself has got to be understood in an impoverished sense here, because our very concept of assertion is tied to the notion of deciding to say something which does or does not mirror what you believe (Williams 1973, p. 146)

In addition to beliefs, the capacity for assertion arguably also requires the capacity to form the right kinds of *intentions*. Ants who fortuitously spell out 'N'djamena is the capital of Chad' in the sand aren't asserting anything, not merely because they fail to understand what they 'say', but also because they aren't *saying* anything—they are not intending to produce speech. According to what we'll call "the attitudes objection", the fact that chatbots lack (the capacity to form) intentions provides a reason to deny that their outputs ever involve genuine acts of assertion (van Woudenberg et al. 2024; Gubelmann, 2024).

### 3.3 The mentalizing objection

Beyond first-order mental states like belief, classical speech act theories hold that assertion involves a number of higher-order, *meta*representational mental states. For instance, Grice's (1957) influential account holds that assertion requires:

(i) an intention to produce a belief in a hearer

(ii) an intention that the hearer recognise this first intention

(iii)    an intention that the hearer forms the belief partly *because* they've recognised this second intention

Other accounts take assertion to require the capacity to keep track of the "common ground" of a conversation (Stalnaker 1974; 1978)—roughly, the set of propositions that are believed by all participants.

Common to these views is the idea that assertion requires mentalizing ('theory of mind') abilities, such as abilities to track, ascribe and reason about the mental states of ones' interlocutors. Thus, a third line of argument against TCA—the mentalizing objection—appeals to the claim that current chatbots lack a theory of mind, or at least that, when producing linguistic outputs, they do not form the sorts of metarepresentational intentions and beliefs required for assertion.

**3.4 The normativity objection**

Assertion is plausibly a norm-governed practice, and many have argued that it is here that the most serious objections to TCA lie (Butlin & Viebahn forthcoming; Arora 2024; Freiman, 2024 Mallory 2023; van Woudenberg et al 2024).

Some norms govern when it is permissible to make an assertion.[2] Others concern the consequences of making an assertion, i.e. what responsibilities or commitments one takes on in asserting *P*. Many make the further claim that these norms are *constitutive* of assertion. While there is much disagreement over what this amounts to (Williamson 2000; Simion & Kelp 2018; García-Carpintero, 2019; Marsili, 2023), most relevant here is the idea that a (putative) speaker is a capable of assertion only if they stand in the *appropriate relation* to its governing norms:

---

[2] Competing accounts variously hold that it's permissible to assert that *P* only if one *knows* that *P;* one *believes* that *P;* one is *epistemically justified* in believing that *P*; or only if *P* is *true* (see Pagin & Marsilli 2021 for an overview of alternative accounts).

> as a game of chess cannot be played unless the rules of chess are in force, so an assertion cannot be made unless the governing norm of assertion is in force and applies to the utterance. (Pagin, 2016, p. 186)

Candidate accounts of what it takes for norms to be in force for a putative speaker include the requirements that the speaker *understands* and/or *intentionally commits* to those norms. Versions of the normativity objection along these lines may, therefore, collapse into the understanding objection and/or the attitudes objection. However, Butlin & Viebahn (forthcoming) defend a different account of what it takes for the norms of assertion to be 'in force', using that account to undermine TCA. They argue that an agent has the capacity to assert only if it can be *sanctioned* for its deviations from the norms of assertion by the agents it interacts with.

There are three conditions on sanctionability (as Butlin and Viebahn understand it).[3] First, it requires that interlocutors keep track of asserters' credibility. Butlin and Viebahn argue that chatbots meet this condition, for we are able to keep track of how reliable a given chatbot is at tracking the truth, giving consistent outputs, and so on.[4] They also argue that some chatbots might meet the second condition on sanctionability, which is that that asserters are sensitive to the fact that interlocutors keep track of their credibility; they point out that during fine tuning processes such as reinforcement learning from human feedback (Ouyang et al., 2022) models' dispositions are shaped by human assessments of their credibility. However they argue that even advanced chatbots fail to meet the third condition on sanctionability—namely, that losing credibility is bad for asserters (it is contrary to their interests). Why does sanctionability require this condition?

---

[3] Butlin & Viebahn also propose a second condition for a system to be capable of assertion: it must produce outputs with descriptive functions (see Section 2). They argue that some language models (those fine-tuned for groundedness/correctness) meet this condition.

[4] Freiman & Miller (2018, p. 431) call this a form "quasi-trust", which they argue "is distinguishable from mere reliance in that it is grounded in normative expectations from its target, as opposed to expectations that are based merely on inductive inference about its reliability".

> Consider a thermometer that has a feedback button that can be pushed in case of inaccurate measurements. Once the button has been pushed a certain number of times, the thermometer enters an automatic routine of self-cleaning and recalibration. This more sophisticated thermometer fulfils [conditions (i) & (ii)] but it still seems implausible to say that the thermometer is producing assertions or that it can be sanctioned for its outputs. (Butlin & Viebahn forthcoming p. 12)

Although Butlin and Viebahn don't advance an account of what having "interests" amounts to, they suggest that "it would likely be necessary either that the system was conscious or that it cared about some properties of itself or its environment" (p. 14), and they argue that there is little reason to think the current generation of chatbots meet either condition.[5] This, they argue, prevents current chatbots from being properly sanctionable and thus, in turn, from having the capacity to make assertions.

## 4. Addressing the Dilemma

Let us take stock. We saw in Section 2 that there is a strong prima-facie case for treating chatbots as conversational partners, capable of making assertions. However, we have also just seen that there are multiple objections to TCA; each of these objections is significant, and their collective force goes a long way towards explaining why few theorists have endorsed TCA.

---

[5] Gubelmann (2024) & Browning (2023) also highlight LLMs' lack of intrinsic interests.

Facing us is a dilemma. Rejecting TCA outright seems unsatisfactory: ideally, any account of chatbots which rejects TCA ought not simply dismiss it, but should show why (although it is in fact false) it seems to be true. Previous rejections of TCA are unsatisfactory in this regard (see Butlin & Viebahn forthcoming; van Woudenberg et al 2024; Gubelmann 2024). Conversely, those who embrace TCA owe us a response to the objections laid out in Section 3. One way of doing this would be along the lines suggested by Cappelen & Dever (Cappelen & Dever, 2021). They suggest that current theories of assertion are likely "parochial" and "too anthropocentric" (p. 135) i.e. too tied to contingent, superficial features of human communication. They predict that if theorists "engage in anthropocentric abstraction … [they will] find some way to create a notion of 'assertion' or 'saying' that can fit ML [machine learning] systems" (p. 135). The idea, in other words, is that the objections rehearsed in Section 3 make demands that a good theory of assertion need not make—or alternatively that, on the best construal of those demands, chatbots already meet them.

Although a full evaluation of this position lies beyond the scope of this paper, we view its prospects of success as dim. The main problem is that it requires first developing a revised "abstracted" theory of assertion, and second, showing that this revised account is superior to existing accounts of assertion with respect to some agreed-upon set of theoretical virtues. Cappelen and Dever do not attempt either task themselves. At any rate, we are not convinced that existing theories of assertion are problematically anthropocentric; nor—for that matter—do we think it likely that the best theory of assertion will turn out to classify current chatbots as fully-fledged asserters. Thus, the second horn of the dilemma remains unattractive.

There is, however, a third possible way of responding to the dilemma. One influential reaction in the literature has involved attempts to 'split-the-difference' between the 'pro' and 'con' cases for TCA. The aim, in other words, is to develop an account that does justice to the motivations for TCA but also takes seriously the objections to it. The remainder of this section considers the two most prominent versions of this response: the *proxy-assertion* view and the *fictionalist* view.

**4.1 Proxy-Assertion**

Perhaps the most influential attempt to 'split the difference' involves treating machine assertion as a case of proxy-assertion, an idea first proposed by Nickel (2013). Nickel suggests that machines can qualify as "speech actants to a substantial degree" (p. 495), but he argues that "ultimate responsibility for artificial speech does not lie with machines, but either with persons or companies, or with nobody at all" (2013, p. 500). His model here is a situation in which a father sends his 8-year-old daughter to buy a bag of flour from the store. As Nickel tells the story, although the daughter speaks, she is not responsible for her speech; instead, that responsibility traces back to her father. It is the father who makes the relevant assertions (or, as the case may be, requests, questions, etc.) and the daughter is involved in those illocutionary acts as a mere proxy (Ludwig, 2018).

By treating chatbots as proxy-asserters, this account promises to evade the dilemma we noted above and do justice to both the 'pro' and 'con' cases. It seems to do justice to 'pro' case, for it recognises that interacting with chatbots does indeed involve genuine assertion. It also seems to do justice to the 'con' case for, by treating chatbots as mere proxies, it avoids the need to show that they can meet the various constraints associated with understanding, mental attitudes, mentalizing and normativity that we identified in Section 3. However, despite its promise, there are two problems with treating chatbots as proxy-asserters—each of which is potentially fatal.

Firstly, in typical cases of proxy assertion there is a clear candidate for the "principal"— the speaker whose assertion is carried out via the proxy. In Nickel's case, we can trace the illocutionary act back to the father. But in typical interactions with an LLM-driven chatbot, there is no clear candidate for the principal. As Freiman (2024, pp. 483–484) notes, what is distinctive about typical interactions with advanced chatbots is that their linguistic outputs are generated algorithmically, with no direct human involvement. This distinguishes them from systems that output canned, prerecorded messages ("Please mind the gap") or device-mediated human testimony (e.g. Siri reading out a text from your brother).

Green & Michel (2022) discuss a case where there seems to be a clearer candidate for the "principal" of AI proxy assertion. Their case involves an AI system deployed by a police department, whose linguistic outputs (e.g. issuing fines for traffic infringements) are implicitly endorsed by the department. While this case may avoid some of the concerns just raised, in so doing, it significantly constrains the scope, and therefore the appeal, of the proxy-assertion response. The considerations for the *pro* case for chatbot assertion were motivated largely by interactions with off-the-shelf general purpose chatbots. But Green & Michel's account, if successful, only grounds chatbot (proxy-)assertion in cases where a chatbot is deployed to generate text on behalf of a person or organisation, and they have adopted an attitude of blanket endorsement towards its outputs.[6]

The second problem with the proxy-assertion account applies even in those narrow cases. As Butlin & Viebahn (forthcoming, pp. 15–18) argue, there are two kinds of cases that might plausibly be called "proxy assertion". In one kind of case, as when someone reads from a script on behalf of someone else, the proxy is not really the one making the assertion—at best they are a conduit for the principal's assertion. In more complex cases, the proxy is granted greater freedom to interpret and speak on behalf of the principal. In this case, the proxy could be seen as making assertions, namely, assertions about what the principal believes. Correspondingly, cases of the second kind bring with them the kinds of mental capacity requirements applicable to ordinary assertions, and the proxy *is* sanctionable if they make unwarranted claims about the principal's beliefs.

---

[6] A parallel points applies to Arora's proxy assertion account (Arora 2024, p. 14). Of course, these authors may see the limited scope of their accounts as a feature, rather than a bug. Nevertheless, they face another challenge, introduced next in the main text.

Advocates of the proxy-assertion account must either hold that chatbots are proxy-asserters in the thin sense, or in the thicker sense. But neither option is attractive. The first horn (seemingly embraced by Arora [2024], whose account also "classifies many trivial machines with a vocal output, such as talking clocks and thermometers, as engaging in proxy assertions" [p. 14]) underplays the ways in which advanced chatbots seem to be more than mere conduits for the speech acts of others—that is, it does not make sufficient contact with the considerations that motivated the "pro" case. But the second horn is equally unattractive, for it fails to address the challenges raised by the "con" case, and faces the burden of showing that chatbots understand what they say, possess the required mental attitudes and mentalizing capacities, and stand in the right relation to the norms of assertion.

**4.2 Fictionalism**

A second attempt to evade the dilemma is advanced by Mallory (2023), who argues that interacting with chatbots involves a kind of ("prop-oriented") make-believe. We are, in effect, treating the chatbots as a fictional character who has a range of linguistic capacities, including the capacity to make assertions.[7]

Like the proxy-assertion view that we've just considered, fictionalism also promises to "split the difference" between the 'pro' and 'con' views, incurring the benefits of both views without the costs. It promises to explain why interacting with chatbots has the phenomenological features that characterize genuine conversation, for episodes of pretending that one is (say) fighting a fire are designed to engender the sense that one is (in some sense) fighting a fire. At the same time, fictionalism doesn't have the commitments that adherence to TCA brings with it. Because the account treats chatbots as a mere 'prop' in a conversational game, it no more presupposes that they have the capacities that are required for literal assertion than treating a cardboard box as a fire-engine requires ascribing genuine fire-fighting capacities to it.

---

[7] Kim's (2023) discussion of chatbots also suggests a fictionalist view.

But for all that, fictionalism fails to provide a satisfactory resolution to the dilemma, for it fails to recognize the differences between advanced LLM-driven chatbots and very simple machines, such as calculators. When it comes to illocutionary capacities, Mallory says, there is no "metaphysically robust" (2023, p. 1097) contrast between chatbots and calculators. Instead, the difference is merely that the former are more effective props in the game of conversational make-believe. But this position is unsatisfying, for there does seem to be a "metaphysically robust" sense in which LLMs are closer to instantiating human-like linguistic agency than a calculator, not just in appearance, but in substance.[8]

We have argued that neither fictionalism nor the proxy-assertion proposal delivers on its promise of 'splitting-the-difference' between the 'pro' and 'con' cases for TCA. In the remaining sections, we develop a third proposal for how to evade the dilemma of chatbot assertion—one that we will argue is more promising than its competitors.

---

[8] Indeed, Mallory himself appears to be moved by the thought that, in the case of modern chatbots, make-believing that that they are linguistic agents affords us knowledge about their "internal workings" (p. 1096). He further suggests that this kind of make-believe, and the knowledge it generates, can be seen as an instance of scientific modelling more broadly. However, this move threatens to undermine the "conservative" (p. 1083) aspirations of the fictionalist account. If all scientific models/theories are fictional in the same sense, then the distinction between *literal* and merely *fictional* asserters loses its grip.

# 5. Proto-Assertion and the Ontogenesis of Assertion

Few capacities have sharp, clearly-defined, boundaries. Think of walking. There is a point at which infants can only crawl, or perhaps walk only with assistance (e.g., an adult holding their hand or a 'walker'). Is a young child who is able to take a few hesitant steps a walker? The issue is moot. They are on their way to becoming a walker—they are in the process of mastering the capacities required for walking—but they are not yet, perhaps, a fully-fledged walker. We might think of them as a 'proto-walker'.

As with walking, so too with speech. There is a period in which the child has the capacity to understand and use a limited range of words, and to deploy those words in the service of illocutionary agency. For example, in response to the question "What did you do today?", a toddler might say, "zoo!". Drawing on your background knowledge, you infer that the toddler went to the zoo. Has the toddler *asserted* that they visited the zoo? One's intuitions might be uncertain. On the one hand it *looks* as though what the toddler is doing is akin to what their older siblings (who are clearly capable of assertion) are doing when they say that they went to the zoo. Indeed, they look to be asserting that they went to the zoo in much the way that by saying "book!" (in a certain context) they are asking for a book to be read to them. Using words to answer questions may not be a paradigm case of assertion (Alston 2000), but it is arguably a step along the path. At the same time, it's unclear whether toddlers meet the various conditions on assertion that we identified in Section 3. For example, it might be doubted whether toddlers understand the semantic content of their utterances. It might be doubted whether they are capable of the range of mental attitudes that are arguably required for assertion. It might be doubted whether they have the mentalizing capacities arguably required for assertion. And it might be doubted whether their speech is norm-guided and sanctionable in the relevant ways.

In fact, the situation is even more complicated than the foregoing suggests, for reflecting on the development of illocutionary capacities reveals that many of the central features of assertion themselves admit of gradations. Young children often have *partial* understanding of the expressions they utter—a toddler might reliably utter "dino!" when pointing at a picture of a dinosaur, but they likely have an impoverished (and perhaps very confused) conception of a dinosaur. Exactly what mentalizing capacities young children have is a matter of ongoing debate (see Butterfill 2020; Carruthers, 2013; Lavelle, 2024; Perner, & Roessler, 2014), but there is little doubt that between early infancy and starting school there are radical changes in a child's capacities to understand, track, and appropriately respond to the mental states of its interlocutors. What about normativity? Rakoczy and Tomasello (2009) present empirical evidence that young children have a "rudimentary grasp" of the norms surrounding assertion (p. 206). In terms of sanctionability, there is a loose sense in which we keep track of young children's adherence to and deviation from the norms of assertion—we might put more faith in their utterances as they get older and prove to be more reliable truth-trackers. Clearly toddlers are not sensitive to their reputation for credibility in the ways that adults are. (There's also no robust sense in which losing credibility is *bad* for young children). But they do seem able to learn from instruction, as when a carer tells a child "that's not a dinosaur, that's a *rhinoceros*".

Where do these considerations leave us? One response is to insist that the challenge here is purely epistemic. Although we may not be able to figure out when children become illocutionary agents, the acquisition of illocutionary capacities has sharp, clearly-defined, boundaries. Before a certain developmental milestone children have no illocutionary capacities at all; following this milestone, they are fully-fledged illocutionary agents. Where exactly that developmental milestone occurs might be difficult to discern (and of course it need not fall at the same point for all children), but—so this line of argument goes—the very nature of illocutionary agency ensures that there must be some such point. In the words of Green and Michel, "we may doubt what clear sense may be attached to something's being a speech act in part, or to some degree. Promising, asserting, appointing and betting are qualitative rather than quantitative notions." (2022, p. 495).

There is, however, another way to think about the acquisition of illocutionary capacities. Rather than assuming that illocutionary agency is defined by a particular developmental milestone, one could think of it in a graded manner in which illocutionary capacities emerge in piecemeal fashion over a period of time. On this view, there is a theoretically important space of 'proto-assertion' through which children pass on their way to becoming full participants in the socially and metacognitively complex practice of assertion. Proto-asserters have either only some of the features characteristic of assertion, and/or those features that they do possess they have only in a limited (partial, faltering) manner. This conception of the ontogenesis of illocutionary capacities seems to us significantly more attractive than the 'all-or-nothing' alternative.[9]

With the concept of proto-assertion in place, let's return to chatbots.

## 6. From Toddlers to Transformers

In Section 2 we noted that a robust case can be mounted for TCA—the claim that chatbots are capable of assertion. Some of their linguistic outputs have the function of tracking the truth; the contrast between their modes seems to correspond to a contrast between illocutionary acts; and they produce (what appear to be) defences and explanations for their putative assertions, thus conforming to the behavioural profile of asserters to a significant degree. Against this, Section 3 reviewed a number of robust objections to TCA—thus generating the dilemma of

---

[9] Note, incidentally, that neither the proxy-assertion nor the fictionalist proposal is at all attractive when it comes to children.

chatbot assertion. We suggest that this dilemma is best resolved by treating chatbots as proto-asserters.[10]

The key point here is that many of the arguments against TCA are, on closer inspection, better construed as simply establishing the *partial* presence of those features.[11]

Consider first the *understanding objection*, according to which current chatbots lack the semantic understanding necessary for assertion. We are sympathetic to the claim that a full grasp of certain concepts requires sensorimotor engagement with the world, but the objection from understanding goes significantly beyond that claim.

For one thing, it's far from evident that sensorimotor capacities are required for grasp of all kinds of concepts. For example, understanding mathematical concepts (ADDITION, INFINITY), moral concepts (WRONG, PERMISSIBLE), and theoretical concepts (ENERGY, GENE) seem to be relatively independent of sensorimotor capacities. What's required, instead, is an adequate sensitivity to their inferential role, and that is something that LLM-driven chatbots may very well have (Piantadosi & Hill, 2022). Even with respect to concepts more closely tied to sensory input (such as RED or LOUD) we should grant that partial understanding might be available via sensitivity to their inferential role.[12] For example, the blind can have partial grasp of RED and the deaf partial grasp of LOUD (Butlin 2023). This is not merely an 'in principle' argument, for research into the internal mechanisms of LLMs trained solely on text has revealed structural correspondences between their activation spaces and a variety of real world structures, suggesting that they may have acquired some degree of semantic understanding, in at least some domains (Abdou et al., 2021; Gurnee & Tegmark, 2023; Li et al., 2023; Nanda, Lee, & Wattenberg, 2023; Yildirim & Paul, 2024). Finally, many of the arguments against understanding in chatbots have been concerned with text-only systems whose inputs, outputs and training data consist solely of text. Thus, objections from the text-bound nature of chatbots evaporate in the face of the fact that most of the leading systems are now "multi-modal"—processing text, audio and video inputs.[13]

---

[10] To pre-empt a potential misunderstanding at the outset, we are not making any of the following claims: that chatbots are *just like* toddlers (clearly, there are innumerable differences), that they are proto-asserters *in precisely the same way* as toddlers (as we'll show, they have a different profile of assertion-relevant features), or that they are *closer* to being full-blooded asserters than toddlers (we remain neutral on this issue, but think it's plausible that toddlers prevail).

[11] In this respect, our approach is convergent with other attempts to capture the "in-between" nature of the capacities of AI systems (Strasser & Schwitzgebel 2024).

[12] See Abreu Zavaleta (2023) for a recent account of the widespread phenomenon of partial understanding.

Although LLMs clearly fall short of a full grasp of the words that they generate (Lake & Murphy 2021), a reasonable case can be made for thinking that they have a *partial* understanding of many concepts—indeed, their grasp of many concepts may be significantly firmer than that of a young child.

What about the *attitudes objection?* Do (current) chatbots lack the propositional attitudes (chiefly, beliefs and intentions) necessary for assertion?

Here we would make two points. The first is that the possession of particular mental attitudes may not be an all-or-nothing matter. Do non-human animals have beliefs and intentions? What about very young children? Do delusions qualify as beliefs? Is self-deception intentional? Rather than restricting ourselves to only two answers to these questions ("Yes they do"; "No they don't"), many would argue that we ought to embrace a variety of "in-between" views, according to which these phenomena involve states that are, although not fully-fledged beliefs or intentions, importantly akin to these states (see e.g. Bayne & Hattiangadi, 2013; Schwitzgebel, 2001; Stich, 1979).

Second, although it is doubtful that chatbots have fully-fledged beliefs or intentions, they may have internal states which exhibit some (if not all) of the functional signature of beliefs. One line of evidence comes from *probing* studies which have found directions in models' internal activations at certain layers that correspond to the truth value of inputs (Azaria & Mitchell, 2023; Burns et al. 2024). Although other theorists have urged caution in interpreting such results (Harding, 2023; Herrmann & Levinstein, 2024; Levinstein & Herrmann forthcoming), even these authors take it as an open empirical question whether LLMs might have belief-like representations.

---

[13] Some conversational agents are also *embodied* and can act on their environment. https://www.figure.ai/

Partly in response to these concerns, more recent studies have employed more careful methodology. Marks & Tegmark (2024) curated datasets to avoid generalisation issues present in previous studies, and also investigated the causal role of activity directions identified by probes. For two models from the LLaMA family, they found components of the models' internal activity (specifically, directions in activation space) which tracked the truth value of inputs, and established through intervention that the these components causally mediated outputs in appropriate ways: shifting activations along the identified directions caused the models to treat false statements as true, and vice-versa.

This is an active area of empirical research, but even from the current evidence, it seems likely that the LLMs underlying advanced chatbots have representations that are at least somewhat belief-like, and that these play a causal role in their production of linguistic outputs. Further, it is noteworthy that even those who have cautioned against naïve ascriptions of beliefs to LLMs treat the concept of belief as involving a cluster of separable properties, each of which may allow for gradations. For instance, Herrmann & Levinstein (2024), who propose four criteria for LLM representations to count as beliefs, stress that "[t]he satisfaction of these requirements come in degrees; in general, the more a representation satisfies these requirements, the more helpful it is to think of the representation as belief-like" (p. 7). This is pertinent to the discussion at hand—chatbots may turn out to be proto-*believers*, which lends further (if indirect) support to treating them as proto-*asserters*.

The question of whether LLMs have *intentions* has received less attention, but we would argue that this possibility should not be dismissed given the situation with respect to belief. More specifically, Todd and colleagues (2023) claim to have found evidence of "function vectors" in LLMs that appear to have a *world-to-mind* direction of fit. These are internal states that seem to serve as compact representations of abstractly-defined input–output functions (or *tasks*), such as translating a text from English to Spanish, identifying the capital city of a country, or finding the antonym of a word. By causally intervening on these vectors, one can trigger or interrupt the performance of a task, in a way that is robust across a wide range of input text, suggesting that these states may play some of the action-guiding functional role of intentions.[14]

What about the capacity of (current) chatbots to engage in the kinds of mentalizing tasks that are arguably required for assertion? Initial investigations suggested optimism about the theory of mind capabilities of LLMs (Kosinski, 2024), but more detailed studies have tended to show that current LLMs' ability to reason about the mental states of others (if that is even the correct framing) is at best brittle (Shapira et al., 2024; Strachan et al., 2024; Ullman, 2023).

However, reflecting on the case of young children suggests that, while this might prevent *full* participation in the practice of assertion, we shouldn't take *proto*-assertion to require the same metarepresentational capacities (which are arguably missing in under-3s) as fully-fledged assertion. Perhaps it requires only some degree of sensitivity to the informational state and requirements of one's audience, which LLMs may indeed have.[15]

---

[14] Hendel and colleagues (2023) present complementary evidence of what they call "task vectors".

[15] It is worth noting that the metarepresentational demands of classical accounts of assertion may be too demanding even for ordinary human adult assertion (Horton, 2018; Westra & Nagel, 2021). Convergent empirical evidence suggests that while cognitively demanding forms of mentalizing are sometimes deployed in communication, we more often rely on cruder, less demanding forms of mentalizing and also on non-metarepresentational heuristics.

What, finally, of worries about *normativity?* One version of the normativity objection was that (on some views) engaging in the practice of assertion requires understanding the relevant norms, and that chatbots lack such understanding. We have already argued against blanket dismissal of understanding in chatbots, and there is reason to think that norms of assertion are not fundamentally beyond the grasp of chatbots; given that their *behaviour* is largely consistent with the norms of assertion (see Section 2), it is plausible that algorithms at least implicitly encode these norms—and there may be independent reasons for requiring no more than implicit understanding of the relevant norms (Simion & Kelp 2018; Williamson 2000, p. 241; cf. Pagin 2016). Most importantly, as argued above, the claim that chatbots are proto-asserters requires only that they have *partial* understanding of the norms governing assertion (indeed, in this respect, they come much closer to the capacities of full-blooded asserters, compared with toddlers).

A different version of the objection from the normativity came from Butlin and Viebahn (forthcoming). As we saw in Section 3.4, they argue that genuine assertion requires sanctionability, and that while (some) chatbots meet the first two conditions on sanctionability, they flout the third. However, in light of the preceding discussion, we suggest that this mixed result is best construed as further evidence that chatbots are *proto*-asserters, rather than for the conclusion that they are straightforwardly incapable of assertion: chatbots might not be sanctionable in the full sense that a human speaker is, but (analogously to children) we may assess them against our norms, and some may be appropriately responsive to correction.

What should we say about Butlin and Viebahn's case of the self-calibrating thermometer (discussed in Section 3.4)? Given that this system also meets the first two conditions on sanctionability, one might worry that the account we have sketched would have to treat it as a proto-asserter and this would trivialise the concept. We share the intuition that such a system would be far less deserving of proto-asserter status than (say) a young child. But (we submit) this is because there are many characteristic features of assertion beyond sanctionability (and descriptive functions) and these are absent in the self-calibrating thermometer. The self-calibrating thermometer (as described by Butlin and Viebahn) is incapable of defending or explaining its would-be assertions; it doesn't switch between apparently different linguistic modes; it doesn't understand its outputs in any robust sense or have abstract belief-like or intention-like representations; and so on. LLM-driven chatbots (and young children) likely possess many of these features to at least some degree. However, if we were to extend Butlin and Viebahn's case by augmenting the self-calibrating thermometer with these features, the intuition that it is not capable of even proto-assertion begins to weaken (as our account would predict).[16]

## 7. Conclusion: At the Edge of Assertion

This paper has advocated for a new conception of chatbots, one which sees them as 'proto-asserters', located somewhere in the space between beings that are fully-fledged asserters and those that are non-asserters. Taking this approach to chatbots, we have suggested, does justice

---

[16] A cousin of the proto-assertion view is the account advanced by Freiman & Miller (2018). They argue that many machines perform what they call "quasi-testimony" (alternatively "quasi-assertion"—they use the terms "testimony" and "assertion" interchangeably [p. 415]), which they define as follows:

> A linguistic output of an instrument or a machine constitutes a quasi-testimony in a given context of use if and only if the machine or instrument has been designed and constructed to produce this output in a manner that sufficiently resembles testimony phenomenologically, and it is in conformity with an epistemic norm that is parasitic on, or sufficiently similar to what is, or would be, an epistemic norm of testimony in the same context. (Freiman & Miller 2018, p. 429)

Although our notion of "proto-assertion" resembles Freiman & Miller's notion of "quasi-testimony", the two concepts do differ. Firstly, their account focuses only on phenomenological resemblance and conformity to basic epistemic norms (e.g. truth-tracking), whereas our notion is hinged to multiple dimensions of full-blooded assertion, including the requirements of sanctionability, mentalizing, understanding and propositional attitudes. Thus, even if Freiman and Miller's notion adequately characterises simple systems such as automated loud speaker announcements or digital timestamps on photographs (both of which they classify as involving "quasi-testimony"), it does not capture what is distinctive about advanced LLM-driven chatbots.

Freiman & Miller also appear to conceive of the logical relationship between testimony and quasi-testimony differently to how we conceive of the logical relationship between assertion and proto-assertion. We take proto-assertion to be distinct from (full-blooded) assertion, standing to it roughly as toddling stands to walking. By contrast, Freiman & Miller seem to suggest that quasi-testimony is a proper *species* of testimony (or assertion), stating that the prefix "quasi" simply "indicates that while instruments assert or testify, they do so differently from humans" (p. 429).

to both the motivations for TCA and the objections to it; further, it doesn't suffer from the problems that face its rivals, the proxy-assertion view and fictionalism.

But perhaps the key contribution of this paper is not the claim that the current generation of chatbots qualify as proto-asserters (which, after all, is hostage to empirical fortune in various ways), but the suggestion that we need to treat the capacity for assertion as graded. The cognitive capacities which underpin assertion are not acquired in one fell swoop, but emerge in piecemeal fashion over a period of time. Whilst these capacities are emerging, children are neither non-asserters nor genuine asserters but proto-asserters. Something similar, we suggest, applies to machines: somewhere in the transition from answer-phones (non-asserters) to the most sophisticated automata of the distant future (fully-fledged asserters) lie proto-asserters: machines that have some of the features of assertion, and/or have those features to varying degrees.

C. (2024). Testing theory of mind in large language models and humans. *Nature Human Behaviour*, *8*(July). https://doi.org/10.1038/s41562-024-01882-z

Strasser, A. & Schwitzgebel, E. (2024). Towards asymmetric joint actions. In A. Strasser (Ed.) *Anna's AI Anthology: How to live with smart machines.* Berlin: Xenomoi Verlag.

Todd, E., Li, M. L., Sharma, A. Sen, Mueller, A., Wallace, B. C., & Bau, D. (2023). *Function Vectors in Large Language Models*. (c), 1–52. Retrieved from http://arxiv.org/abs/2310.15213

Ullman, T. (2023). *Large Language Models Fail on Trivial Alterations to Theory-of-Mind Tasks*. Retrieved from http://arxiv.org/abs/2302.08399

van Woudenberg, R., Ranalli, C., & Bracker, D. (2024). Authorship and ChatGPT: a Conservative View. *Philosophy and Technology*, *37*(1), 1–26. https://doi.org/10.1007/s13347-024-00715-1

Westra, E., & Nagel, J. (2021). Mindreading in conversation. *Cognition*, *210*(January). https://doi.org/10.1016/j.cognition.2021.104618

Williams, B. (1973). *Problems of the Self*. Cambridge: Cambridge University Press.

Williamson, T. (2000). *Knowledge and Its Limits*. Oxford: Oxford University Press.

Wright, C. (1992). *Truth and objectivity.* Harvard University Press.

Yildirim, I., & Paul, L. A. (2024). From task structures to world models : what do LLMs know ? *Trends in Cognitive Sciences*, 1–12. https://doi.org/10.1016/j.tics.2024.02.00829